\documentclass{article}
\usepackage{spconf,amsmath,graphicx}
\usepackage{booktabs}

\usepackage{color}
\newcommand{\dice}{\mathit{dice}}
\newcommand{\xu}[1]{#1}
\newcommand{\attention}[1]{{#1}}

\title{White matter hyperintensity segmentation from T1 and FLAIR images using fully convolutional neural networks enhanced with residual connections}
\name{Dakai Jin, Ziyue Xu, Adam P. Harrison, Daniel J. Mollura \thanks{Corresponding author: ziyue.xu@nih.gov. This research is supported by CIDI, the intramural research program of the National Institute of Allergy and Infectious Diseases.}}
\address{Center for Infectious Disease Imaging, \\ Department of Radiology and Imaging Sciences, \\ National Institutes of Health (NIH), Bethesda, MD 20892, USA.}

\begin{document}
\maketitle

\begin{abstract}
Segmentation and quantification of white matter hyperintensities (WMHs) are of great importance in studying and understanding various neurological and geriatric disorders. Although automatic methods have been proposed for WMH segmentation on magnetic resonance imaging (MRI), manual corrections are often necessary to achieve clinically practical results. Major challenges for WMH segmentation stem from their inhomogeneous MRI intensities, random location and size distributions, and MRI noise. The presence of other brain anatomies or diseases with enhanced intensities adds further difficulties. To cope with these challenges, we present a specifically designed fully convolutional neural network (FCN) with residual connections to segment WMHs by using combined T1 and fluid-attenuated inversion recovery (FLAIR) images. Our customized FCN is designed to be straightforward and generalizable, providing efficient end-to-end training due to its enhanced information propagation. We tested our method on the open WMH Segmentation Challenge MICCAI2017 dataset, and, despite our method's relative simplicity, results show that it performs amongst the leading techniques across five metrics. More importantly, our method achieves the best score for hausdorff distance and average volume difference in testing datasets from two MRI scanners that were not included in training, demonstrating better generalization ability of our proposed method over its competitors.
\end{abstract}


\section{Introduction}
\label{sec:introduction}

\begin{figure}[htb]
\centering
\centerline{\includegraphics[width=8.5cm]{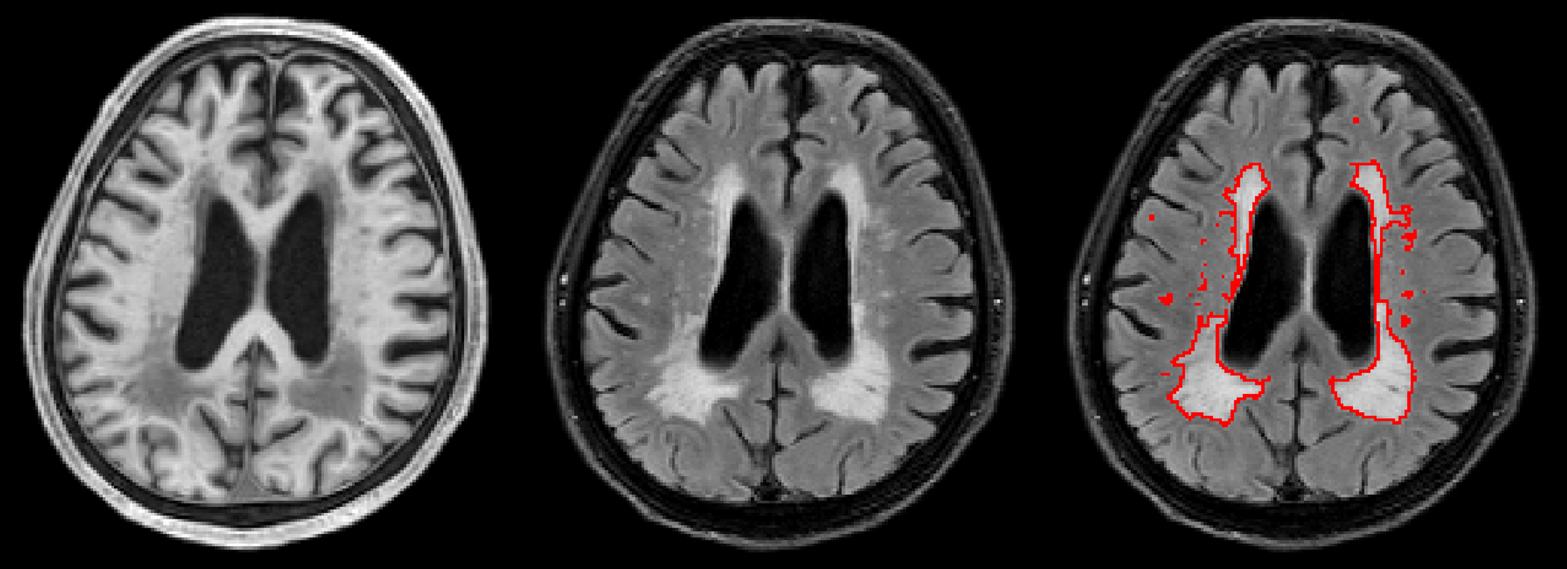}}
\caption{\small \attention{Example slices of WMHs on T1 image with decreased intensities (left) and FLAIR image with enhanced intensities (middle).} The boundary of ground truth WMH is indicated in red overlapped in FLAIR images (right). Note the random locations, variant sizes and inhomogeneous MR intensities of WMHs.}
\label{fig:2Ddemo}
\end{figure}

White matter hyperintensities (WMHs) are brain areas in the cerebral white matter with increased signal intensity on T2-weighted or fluid-attenuated inversion recovery (FLAIR) magnetic resonance imaging (MRI) scans. \xu{WMHs} are commonly found in the brain of healthy elders and patients diagnosed with small vessel disease and other neurological disorders~\cite{wardlaw2013neuroimaging}.  Accurate quantification of WMH volume, location and shape is of great \xu{importance} for tracking disease progression and evaluating treatment effects. Thus, a reliable and efficient WMH segmentation and quantification method is highly \xu{desirable}. Due to the \xu{labor-intensive} nature and high intra- and inter- observer variability of WMH manual delineation, several automated, and non deep learning, methods have been \xu {proposed} in the literature~\cite{garcia2013review, caligiuri2015automatic}. However, none of the methods achieves \xu{reliable performance close to human readers}. Major challenges for WMH segmentation include: \xu{noise and imaging artifacts, inhomogeneous intensities, random locations, and variabilities in sizes} (see Fig.~\ref{fig:2Ddemo}). The presence of other brain anatomies or diseases with enhanced intensities on FLAIR image adds further difficulties.

Deep neural networks have achieved great success in both natural and medical image domains~\cite{Simonyan2014, He2016, Ronneberger2015}. \xu{Specifically for object segmentation,} the fully convolutional neural network (FCN)~\cite{long2015fully} architecture is an efficient option, with U-Net~\cite{Ronneberger2015} prominently succeeding in segmenting finer scale objects by integrating shortcuts from stages in the downsampling path to the corresponding stage in the upsampling path. \xu{To further promote information propagation within a network,} deep residual network (ResNets)~\cite{He2016, wu2016wider} have been shown to be effective. \xu{This can} be quite beneficial for training deep networks with limited data in the medical imaging domain. \xu{Deep learning methods have also been proposed} for WMH segmentation. For instance, Ghafoorian et al.~\cite{ghafoorian2017location} have recently reported different fusion strategies to segment WMHs in a patch-based manner. However, their method has not been comprehensively evaluated and patch-based object segmentation is often not as efficient as FCNs. \attention{Recently, the WMH Segmentation Challenge at MICCAI 2017 also featured many deep-learning solutions~\cite{WMHChallenge}, many of which were heavily designed for the challenge in question.}

In this paper, we present a customized U-Net FCN with residual connections, \attention{namely \textit{ResU-Net}}, to segment WHM by using combined T1 and FLAIR images.  \attention{Given the high sensitivity of challenge rankings to metric weights~\cite{fishbaugh2017data}, we put less emphasis on achieving the highest rank in the WMH Segmentation Challenge. Instead, our goal is to perform comparably to the Challenge leaders while demonstrating excellent generalization ability, thereby providing confidence that our solution can be effectively used outside the confines of the challenges.} We demonstrate that the additional residual connections help to capture more fine-scale WMHs while reducing false positives. Furthermore, combining FLAIR and T1 images helps the network learn a more robust WMH segmentation. \attention{Importantly, our solution remains straightforward, performing amongst the leading techniques of the WMH Segmentation Challenge with leading performance on datasets not included in the training regimen.}

\section{Methods}
\label{sec:methods}
\begin{figure}[htbp]
\centering
\centerline{\includegraphics[width=8.5cm]{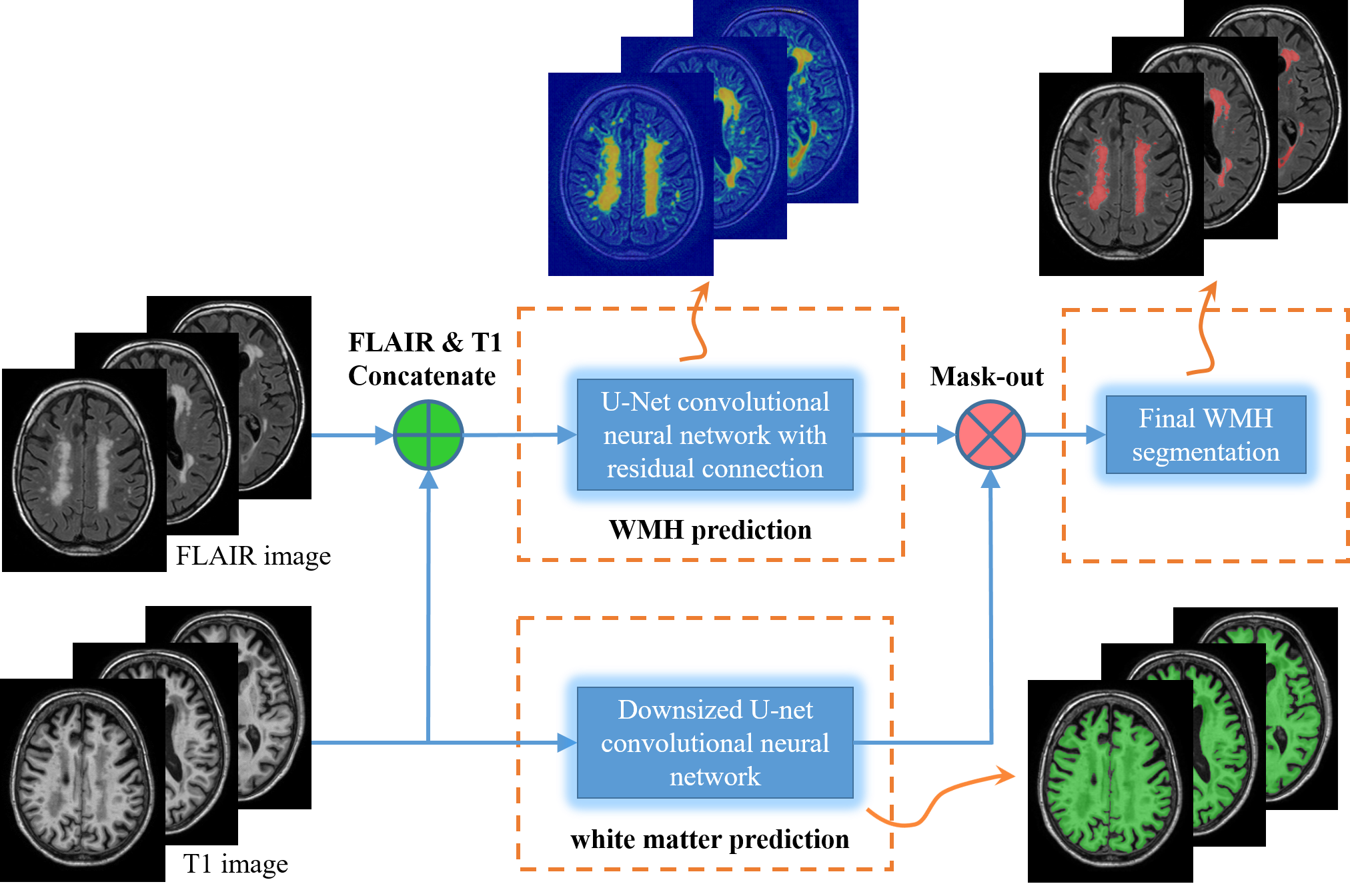}}
\caption{\small Work flow of the proposed WMH segmentation method.}
\label{fig:workflow}
\end{figure}

The proposed WMH segmentation algorithm is illustrated in Fig.~\ref{fig:workflow}. Two major steps are involved: (1) rough white matter segmentation from T1 images by a trimmed U-Net FCN with further morphological refinement; and (2) WMH segmentation using proposed \attention{ResU-Net} on combined T1 and FLAIR sequence images. 

\subsection{\xu{White matter segmentation}}
Since WMHs only appear inside the white matter region, it would be \xu{helpful to constrain the WMH searching area}. Hence, we train a simple downsized 2D U-net FCN, which removes the last pooling operations in original U-Net. Other parameters, such as pooling, up-convolution, feature channels are set the same as the original U-Net~\cite{Ronneberger2015}. As a further refinement, we keep the largest connected-component and apply a morphological dilation to ensure we obtain complete coverage of the white matter area. This white matter mask is used to confine the regions for WMH segmentation in the next step.

\subsection{WMH segmentation:}
\noindent\textbf{U-Net architecture with residual connections (ResU-Net):}
Deep residual network (ResNets) have demonstrated promising results in many benchmarks in computer vision area. Its key component, residual learning, is implemented through a "shortcut connection" that bypasses the nonlinear layers with an identity mapping. In this way, it recasts the nonlinear layers to fit a residual function with respect to its input, which is demonstrated easier to be optimized by the stochastic gradient descent (SGD) method \cite{He2016}. ResNets consist of many stacked "residual blocks", each of which can be expressed as:
\begin{align}
	x_{\ell+1}=x_{\ell} + \mathcal{F}(x_\ell, \mathcal{W}_\ell) \mathrm{,}
\end{align}
where $x_{\ell}$, $x_{\ell+1}$ and $\mathcal{W}_{\ell}$ are the input, output and the convolutional weights of the $\ell^{th}$ residual block, respectively, and $\mathcal{F}(·)$ denotes the residual function corresponding to the $\ell^{th}$ unit.

Recently, ResNets have been shown to behave like an ensemble of sub-networks with different layer depths, therefore, it not only promotes the signal and gradient propagation within a network, but also implicitly alleviates the problem of over-fitting \cite{wu2016wider, veit2016residual}. 

\begin{figure*}[htbp]
\centering
\centerline{\includegraphics[width=16cm]{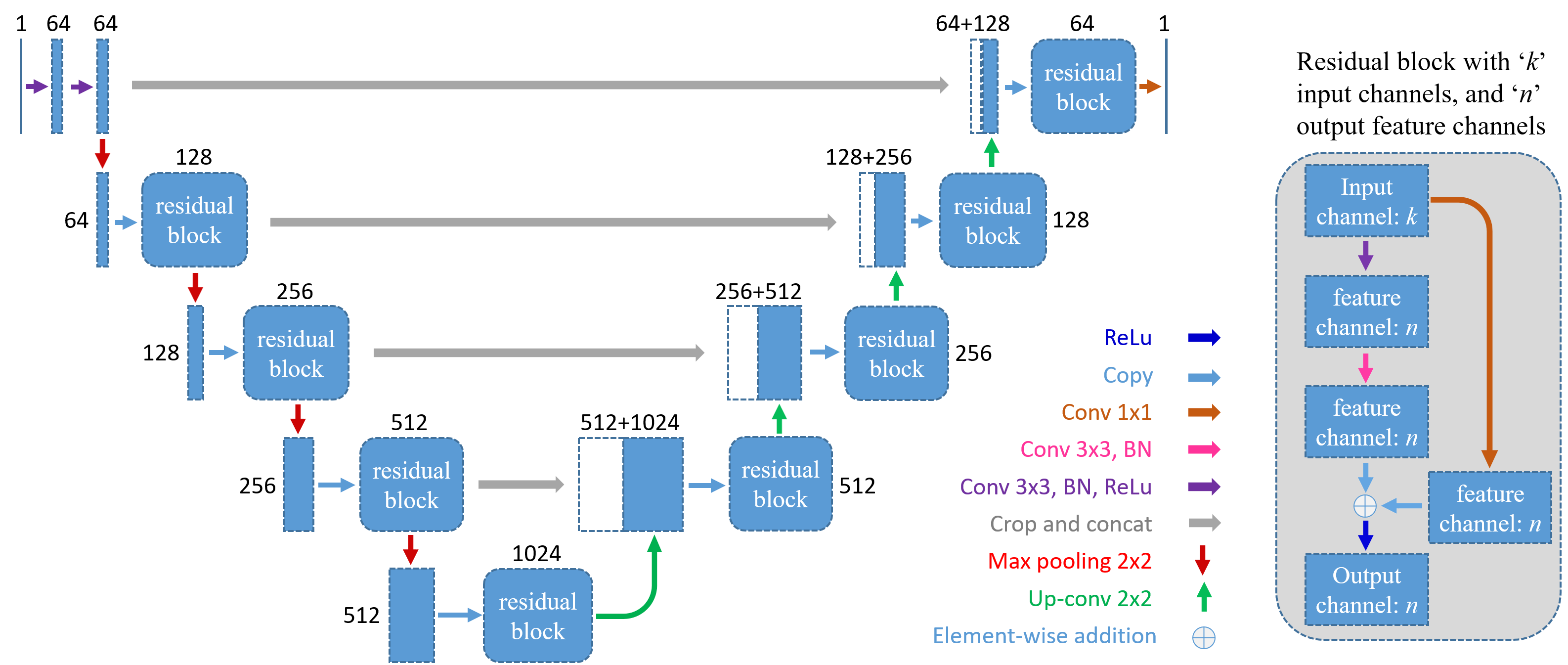}}
\caption{\small Proposed ResU-Net architecture for WMH segmentation. Each box corresponds to a multi-channel feature map with channels number denoted on the top or side of the box. Arrows represent different operations. Operations inside a residual block is illustrated.}
\label{fig:unet_resent}
\end{figure*}

To take advantage of these benefits, we integrate residual connections within a U-net architecture. See Fig.~\ref{fig:unet_resent} for a graphical illustration. The major network frame still consists of a fine-to-coarse downsampling path and a coarse-to-fine upsampling path with shortcut connections to better capture finer scale details. 
\attention{For every two convolutional layers at the same resolution stage in U-Net, we convert them into a residual block.} Unlike the original ResNet, $1\times1$ convolution operation is needed for every residual block in the U-Net architecture to match the input and output number of feature channels.   Other network parameters are the same as the original U-Net.

\noindent\textbf{Global weighted loss function:}
In our end-to-end training, the loss function is computed over all pixels in a training slice. Due to the sparsity of WMHs, the distribution of WMHs and background pixels is highly biased. Therefore, a global class-balancing weight is applied in the loss function as follows:
\begin{align}
	loss=-\beta\sum_{j\in Y_{+}}\log{\hat{y}_{j}}-(1-\beta)\sum_{j\in Y_{-}}\log\left(1-\hat{y}_{j}\right) \mathrm{,}
\end{align}
where $\hat{y}_{j}$ is the computed value after the final convolutional layer, $Y_{+}$ and $Y_{-}$ represent the set of the foreground and background WMH labels, respectively; and $\beta=\mathit{mean}\left(|Y_{-}|/|Y|\right)$  is a global weight pre-computed over the entire training data.

\section{Experiments and Results}
\label{sec:results}
\noindent\textbf{Data and preprocessing:} 
We used the dataset from the WMH Segmentation Challenge MICCAI 2017~\cite{WMHChallenge} for training and testing, where images were acquired from five different scanners of three different vendors in three different hospitals. For each subject, biased field corrected 3D T1 and FLAIR images are provided, which have also been registered. A total of 60 images from 3 scanners are provided by the challenge for training, while the other 110 images from all 5 scanners are hold out for testing. The WMH manual reference standard is generated and verified according to the criterion provided in~\cite{wardlaw2013neuroimaging}.  \attention{For the white matter segmentation, due to the fact that FSL software ~\cite{Jenkinson12} could sometimes generate completely failed white matter masks, we chose to train a downsized FCN instead of using FSL software, where the training label is derived from the successful cases of FSL.}
% * <adam.p.harrison@gmail.com> 2017-10-23T15:08:20.859Z:
% 
% > We used the FSL software ~\cite{Jenkinson12} to get WM masks for training the WM segmentation FCN. 
% Why not just use the FSL for the mask themselves then?
% 
% ^.

For both white matter and WMH FCN, we train slice-by-slice in the axial direction, augmenting with random rotations and flips. For the WMH data, both T1 and FLAIR images have been normalized to the range of $[0, 1]$ using the minimum and maximum values from the white matter mask computed by the white matter FCN. We randomly selected $15\%$ of images for validation, resulting in $\sim50k$ training samples for WMH FCN. Based on the training set distribution, we set the global loss weight to $0.025$. Training stopped after $4$ epochs.
% * <adam.p.harrison@gmail.com> 2017-10-23T15:11:39.765Z:
% 
% > {22.5, 45, 67.5, 90, 112.5, 135 and 157.5 degrees counter-clockwise}
% This seems like an unrealistic data augmentation. 
% 
% ^ <dakai.jin@gmail.com> 2017-10-23T15:55:08.416Z.

% \noindent\textbf{WMH FCN optimization:}
% When training WMH FCN, \xu{based on the training data statistics,} the global weight in loss function is set to the value of 0.025 to balancing the WMH and non-WMH pixels. Optimization was performed using stochastic gradient descent, with initial learning rates of 1e-5, a momentum of 0.99, and a weight decay of 0.05. Learning rate is decreased by 10 after training 2 epochs, and we stop the training after 4 epochs. 

\noindent\textbf{Quantitative evaluation metrics:}
Challenge organizers provided $5$ quantitative metrics for evaluation: dice coefficient, hausdorff distance in ``mm'' unit  (modified as 95th percentile) (H95), average volume difference\% (AVD\%), sensitivity (recall) and F-1 score for individual lesions. Each metric is averaged over all test scans, and each team's ranking is given in the range of $[0,1]$, with the value computed relative to the range of best and worse performance of competitors. For example for dice metric, team-k's dice rank is: $1-(\dice_k-\dice_{min})/(\dice_{max}-\dice_{min}) $. Finally, five ranks are averaged into an overall rank. A lower rank is better.

\begin{figure}[htbp]
\centering
\centerline{\includegraphics[width=8.5cm]{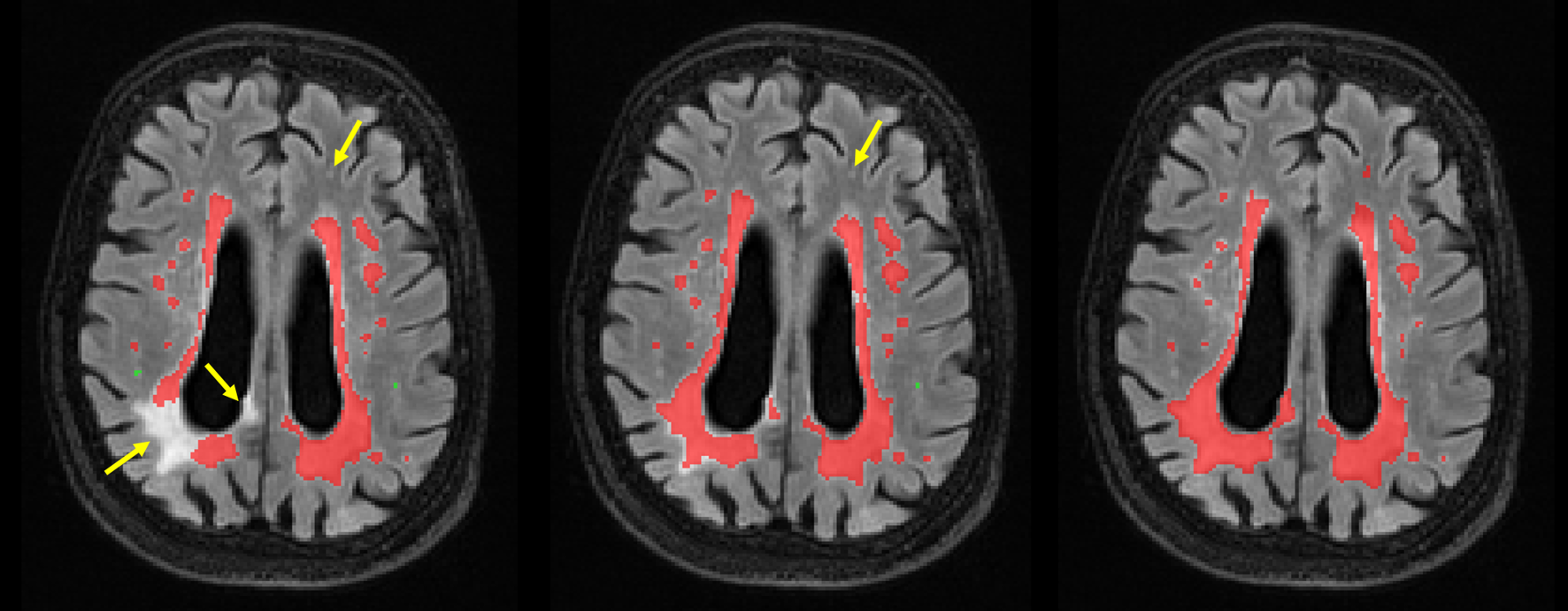}}
\caption{\small Qualitative comparison of WMH segmentation of standard U-Net (left), proposed ResU-net (middle) and manual reference (right) from one validation data. Green dots represent false positives and yellow arrows indicate false negatives.}
\label{fig:quality_result}
\end{figure}

\begin{table}[htbp]
\small
\caption{Quantitative results for the top 5 teams in WMH Segmentation Challenge on all testing data from five scanners. Note that for H95 and AVD\%, lower values are better.}
\label{tab:overallRank}
\centering
    \begin{tabular}{c|c|c|c|c|c|c}

        team & rank & Dice & H95 & AVD\% & Recall & F-1\\
        \hline
        sysu\_media & 0.008 & \textbf{0.80} & \textbf{6.3} & 21.9 & \textbf{0.84} & 0.76 \\
        \hline
        cain  & 0.037 & 0.78 & 6.8 & 21.7 & 0.83 & 0.70 \\
        \hline
        nlp\_logix & 0.049 & 0.77 & 7.2 & \textbf{18.4} & 0.73 & \textbf{0.78} \\
        \hline
        \textbf{nih\_cidi\_2} & 0.060 & 0.75 & 7.35 & 27.26 & 0.81 & 0.69 \\
        \hline
        nic-vicorob & 0.074 & 0.77 & 8.3 & 28.5 & 0.75 & 0.71 \\
  \end{tabular}
\end{table}

\noindent\textbf{Comparison with standard U-Net:}
To measure the impact of the residual connections, we also trained a \xu{standard U-Net} using the same training data and optimization parameters. The qualitative comparison between plain U-Net and the proposed ResU-Net is shown in Fig.~\ref{fig:quality_result}.  Note that proposed ResU-Net captures the WMHs well overall with only one false positive (green dots) and one false negative (yellow arrow) as compared to manual references. In contrast, standard U-Net generates two false positives (green dots) and clearly missed two true lesions (yellow arrow).  Quantitative results on our validation dataset reveal roughly a $\sim$3\% increase in the dice coefficient and F-1 scores. Both qualitative and quantitative results show that residual connections in the U-Net architecture are effective in segmenting WMHs.

\noindent\textbf{Results on WMH Segmentation Challenge MICCAI 2017:}
Our proposed method, named \textbf{nih\_cidi\_2}, achieves 4th place out of $22$ participating submissions in the overall $110$ image testing data\footnote{Results can be found on WMH Segmentation Challenge MICCAI 2017 website: http://wmh.isi.uu.nl/results/}. \attention{However, it is important to point out that the top 5 teams outperformed other teams with a clear margin (average $\geq$6\% in dice, recall and F-1 scores), while they themselves perform within marginal difference (Table \ref{tab:overallRank}).} The specific ranking place may not be that important in this situation, as the commonly used average or accumulated ranking scheme may not reflect the importance of difference metrics for specific problems; hence, a different weighting scheme can dramatically change the ranking on real anatomical data~\cite{fishbaugh2017data}.

\attention{A more detailed look at the results reveals that our proposed method actually performs comparatively better on the testing data from 2 MRI scanners that were not included in training. We summarized the quantitative results for these 2 ``unseen'' scanners in Table \ref{tab:qualiEva}}. Using the same ranking scheme, our method now performs the 2nd best among the top 5 methods, and achieves the best H95 and AVD\% scores. \attention{Although we do not achieve the best rank, however, our method only has a clear drop in dice metric, while all other 4 teams (including top performing team \textbf{sysu\_media}) have a clear drop in at least $3$ metrics. For clear drop of a metric, we refer to a $\geq$3\% decrease in dice, recall and F-1 scores or $\geq$50\% increase in H95 and AVD\%.}  Among the $5$ best performing teams, \textbf{nlp\_logix} and \textbf{nic-vicorob} used patch-based prediction combined with location fusions and cascades of 3D neural networks, respectively, both of which are computationally expensive and complex. \textbf{cian} applied multi-dimensional gated recurrent units, which often requires sophisticated optimization schemes. Not surprisingly, our method outperforms these 3 methods on data from unseen scanners, attesting to the benefit of using a more straightforward and less customized solution. Like us, the top performing team, \textbf{sysu\_media}, used a standard U-Net architecture, but customized its kernel size and number for the challenge. Additionally, \textbf{sysu\_media} applied $3$-classifier ensembling, which we did not. As well, they applied a dataset-dependent post-processing that removed a chosen number of starting and ending slices to reduce false positives. This approach may not generalize beyond the challenge datasets. \attention{In contrast, our method remains generic, meeting our goal of achieving high performance with the aim of deployment beyond the challenge setting.}

\begin{table}[t]
\small
\caption{Quantitative results for the top 5 teams in WMH Segmentation Challenge on testing data from GE1.5T and PETMR scanners that were not included in training.}
\label{tab:qualiEva}
\centering
    \begin{tabular}{c|c|c|c|c|c|c}

        team & rank & Dice & H95 & AVD\% & Recall & F-1\\
        \hline
        sysu\_media & 0.008 & \textbf{0.74} & 11.0 & 26.2 & \textbf{0.87} & 0.72 \\
        \hline
        \textbf{nih\_cidi\_2}  & 0.043 & 0.70 & \textbf{9.7} & \textbf{21.9} & 0.79 & 0.68 \\
        \hline
        cain & 0.053 & \textbf{0.74} & 14.1 & 28.4 & 0.82 & 0.66 \\
        \hline
        nic-vicorob & 0.077 & 0.71 & 13.5 & 56.3 & 0.81 & 0.62 \\
        \hline
        nlp\_logix & 0.082 & 0.68 & 13.0 & 27.9 & 0.66 & \textbf{0.73} \\
  \end{tabular}
\end{table}

\attention{Overall, as compared to other top methods with similar performance, our method is straightforward, generic, computationally efficient, and generalizes better to data acquired from unseen scanners, which well fits the need in practical application.}

\noindent\textbf{Potential improvement:} A closer look at our computed white matter masks shows that they sometimes miss a significant part of white matter regions close to the posterior and inferior regions of the brain, where may well include WMHs. This is due to some error-prune white matter references generated by FSL software when training white matter FCN. Because of this, the already detected WMHs by our ResU-Net are falsely removed, and the performance (especially dice metric) are artificially decreased. Therefore, we believe a better white matter mask from well-labeled training data will further improve our WMH segmentation results.

\section{Conclusion}
In this study, we presented a simple and robust WMH segmentation method by using the proposed ResU-Net. Evaluated on the open WMH Segmentation Challenge at MICCAI 2017, results show that the residual connection does provide considerable improvement in segmenting WMHs, and our method performs among the top leading methods. Our method also achieves the best score for H95 and AVD\% in testing datasets from the two ``unseen'' MRI scanners, further demonstrating the good generalizbility that was our main aim. Thus, we demonstrate that with a straightforward, yet well-thought out architecture, one can achieve excellent WMH segmentation performance that remains generalizable for situations outside any specific challenge setting.

\label{sec:ref}
\bibliographystyle{IEEEbib}
\bibliography{refs}

\begin{thebibliography}{10}

\bibitem{wardlaw2013neuroimaging}
Joanna~M Wardlaw, Eric~E Smith, Geert~J Biessels, Charlotte Cordonnier, Franz
  Fazekas, Richard Frayne, Richard~I Lindley, John T~O'Brien, Frederik Barkhof,
  Oscar~R Benavente, et~al.,
\newblock ``Neuroimaging standards for research into small vessel disease and
  its contribution to ageing and neurodegeneration,''
\newblock {\em The Lancet Neurology}, vol. 12, no. 8, pp. 822--838, 2013.

\bibitem{garcia2013review}
Daniel Garc{\'\i}a-Lorenzo, Simon Francis, Sridar Narayanan, Douglas~L Arnold,
  and D~Louis Collins,
\newblock ``Review of automatic segmentation methods of multiple sclerosis
  white matter lesions on conventional magnetic resonance imaging,''
\newblock {\em Medical image analysis}, vol. 17, no. 1, pp. 1--18, 2013.

\bibitem{caligiuri2015automatic}
Maria~Eugenia Caligiuri, Paolo Perrotta, Antonio Augimeri, Federico Rocca, Aldo
  Quattrone, and Andrea Cherubini,
\newblock ``Automatic detection of white matter hyperintensities in healthy
  aging and pathology using magnetic resonance imaging: A review,''
\newblock {\em Neuroinformatics}, vol. 13, no. 3, pp. 261--276, 2015.

\bibitem{Simonyan2014}
K.~Simonyan and A.~Zisserman,
\newblock ``Very deep convolutional networks for large-scale image
  recognition,''
\newblock {\em CoRR}, vol. abs/1409.1556, 2014.

\bibitem{He2016}
Kaiming He, Xiangyu Zhang, Shaoqing Ren, and Jian Sun,
\newblock ``Deep residual learning for image recognition,''
\newblock in {\em IEEE Conference on Computer Vision and Pattern Recognition},
  June 2016.

\bibitem{Ronneberger2015}
Olaf Ronneberger, Philipp Fischer, and Thomas Brox,
\newblock {\em U-Net: Convolutional Networks for Biomedical Image
  Segmentation}, pp. 234--241,
\newblock Springer International Publishing, 2015.

\bibitem{long2015fully}
Jonathan Long, Evan Shelhamer, and Trevor Darrell,
\newblock ``Fully convolutional networks for semantic segmentation,''
\newblock in {\em Proceedings of the IEEE Conference on Computer Vision and
  Pattern Recognition}, 2015, pp. 3431--3440.

\bibitem{wu2016wider}
Zifeng Wu, Chunhua Shen, and Anton van~den Hengel,
\newblock ``Wider or deeper: Revisiting the resnet model for visual
  recognition,''
\newblock {\em arXiv preprint arXiv:1611.10080}, 2016.

\bibitem{ghafoorian2017location}
Mohsen Ghafoorian, Nico Karssemeijer, Tom Heskes, Inge~WM van Uden, Clara~I
  Sanchez, Geert Litjens, Frank-Erik de~Leeuw, Bram van Ginneken, Elena
  Marchiori, and Bram Platel,
\newblock ``Location sensitive deep convolutional neural networks for
  segmentation of white matter hyperintensities,''
\newblock {\em Scientific Reports}, vol. 7, 2017.

\bibitem{WMHChallenge}
``{WMH Segmentation Challenge at MICCAI 2017},'' http://wmh.isi.uu.nl/.

\bibitem{fishbaugh2017data}
James Fishbaugh, Marcel Prastawa, Bo~Wang, Patrick Reynolds, Stephen Aylward,
  and Guido Gerig,
\newblock ``Data-driven rank aggregation with application to grand
  challenges,''
\newblock in {\em International Conference on Medical Image Computing and
  Computer-Assisted Intervention}. Springer, 2017, pp. 754--762.

\bibitem{veit2016residual}
Andreas Veit, Michael~J Wilber, and Serge Belongie,
\newblock ``Residual networks behave like ensembles of relatively shallow
  networks,''
\newblock in {\em Advances in Neural Information Processing Systems}, 2016, pp.
  550--558.

\bibitem{Jenkinson12}
Mark Jenkinson, Christian F.Beckmann, Timothy~E.J. Behrens, Mark W.Woolrich,
  and Stephen~M. Smith,
\newblock ``Fsl,''
\newblock {\em NeuroImage}, vol. 62, no. 2, pp. 782--790, 2012.

\end{thebibliography}

\end{document}